\documentclass[11pt]{article}

\usepackage[final]{acl}

\usepackage{times}
\usepackage{latexsym}
\usepackage{textcomp}
\usepackage{pifont}
\usepackage{natbib}
\usepackage{adjustbox}
\usepackage{graphicx}
\usepackage{subcaption}
\usepackage{algorithm}
\usepackage{algorithmic}
\usepackage{xcolor}

\usepackage{multicol}
\usepackage{enumitem}
\setlist[itemize]{leftmargin=0.5cm,labelsep=0.15cm}

\usepackage[utf8]{inputenc}
\usepackage[T1]{fontenc}

\usepackage{xspace}
\newcommand{\eb}{ERNIE 4.0\xspace}

\usepackage{CJKutf8}
\newcommand{\Chi}[2]{%
  \csname CJK*\endcsname{UTF8}{gbsn}%
    \CJKchar{"#1}{"#2}%
  \csname endCJK*\endcsname
}

\usepackage{amsmath}
\usepackage{booktabs}

\usepackage[T1]{fontenc}

\usepackage[utf8]{inputenc}

\usepackage{microtype}

\usepackage{inconsolata}

\usepackage{graphicx}

\newlength{\bulletindent}
\title{DuIVRS-2: An LLM-based Interactive Voice Response System for Large-scale POI Attribute Acquisition}

\author{Le Zhang\thanks{\ \ These authors contributed equally to this work.},  Shengming Zhang\footnotemark[1],  Rui Zha\footnotemark[1],  Yunpeng Wu, { Jingbo Zhou\thanks{\ \ Project lead and corresponding authors.},  Jizhou Huang\footnotemark[2]} \\
  Baidu Inc., Beijing, China \\
  \texttt{\{zhangle0202, michaelzhangshengming, crui0210\}@gmail.com, wuyunpeng@baidu.com}\\
   \texttt{zhoujingbo@baidu.com,  huangjizhou01@baidu.com}
}

\begin{document}
\maketitle
\begin{abstract}
Accurate Point of Interest (POI) attribute acquisition is essential for location-based services, yet traditional modular Interactive Voice Response (IVR) systems suffer from error accumulation and high maintenance overhead. We present DuIVRS-2, a large language model (LLM)-based end-to-end framework designed for large-scale POI attribute acquisition at Baidu Maps. To address the long-tail distribution of real-world interactions, our methodology first employs a finite state machine (FSM)-guided data augmentation strategy to synthesize a balanced and diverse training dataset. We then streamline dialogue management via a selective generation scheme combined with a Chain-of-Thought (CoT) mechanism, which ensures output stability and effectively eliminates hallucinations in industrial settings. To facilitate continuous policy refinement with minimal manual effort, we design a cooperative iterative learning framework that leverages a dual-evaluator voting system. Deployed in production for two months, DuIVRS-2 processed 0.4 million calls daily and achieved a 83.9\% Task Success Rate (TSR), outperforming its predecessor by 4 percentage points while maintaining a low reaction time of 130ms. This work provides a production-proven reference for developing robust, cost-effective LLM agents for large-scale industrial dialogue applications.

\end{abstract}

\section{Introduction}

In commercial map applications like Baidu Maps, users are provided with a wide range of Point of Interest (POI) information including attributes like the POI's name and address. 
This rich information empowers various POI-related tasks in Baidu Maps, such as POI retrieval~\cite{huang2021hgamn}, POI recommendation~\cite{chen2021curriculum, fan2021meta}, geography-language pre-training~\cite{huang2022ernie}, and intelligent voice assistant~\cite{huang2022duiva}.
However, the landscape of POI data is both highly dynamic and vast in scale, with recent statistics indicating that 74.5\% of the POIs on Baidu Maps were updated in 2020. Given the sheer volume of POIs, manually acquiring attributes for hundreds of millions of them is impractical due to the labor-intensive and time-consuming nature of the task. This emphasizes the critical need for accurate and current POI attributes to enhance user experience and service effectiveness in practical applications.
Consequently, POI attribute acquisition, which focuses on filling in missing attributes or revising the current values of a POI, plays an indispensable role in enhancing the quality of service and user satisfaction in map applications.

\begin{figure}[t]
\includegraphics[width=0.8\linewidth,trim={0.0cm 0.1cm 0.0cm 0.0cm},clip]{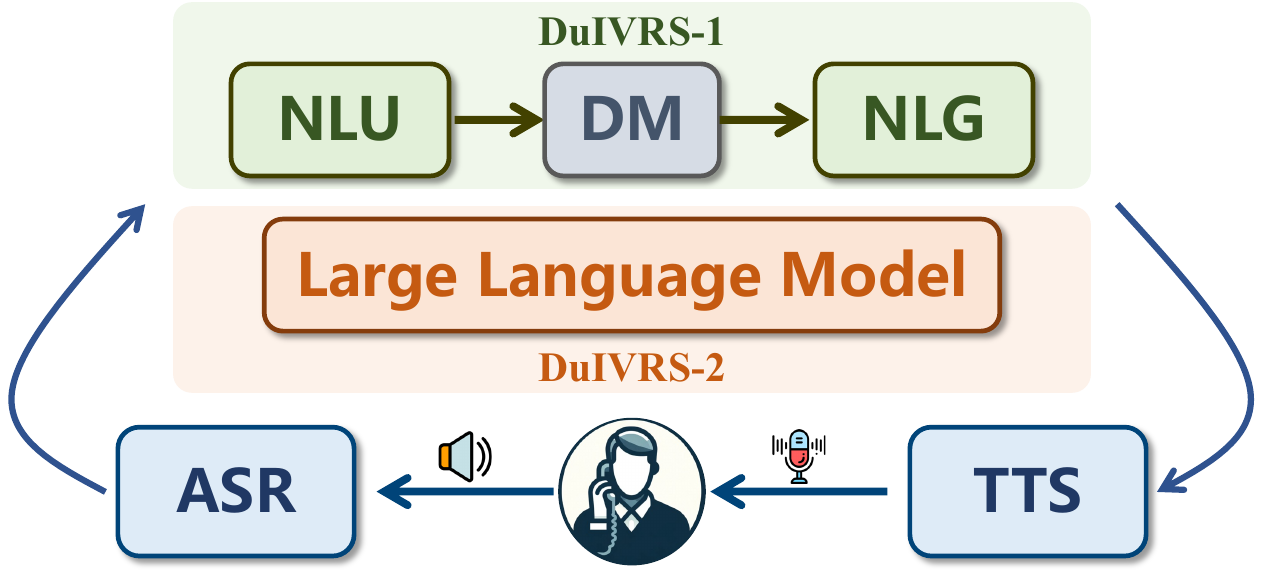} 
% \vspace{-2mm}
\centering
\caption{Comparison of DuIVRS-1 and DuIVRS-2.}
\label{duivrs}
\vspace{-5mm}
\end{figure}

Prior studies have explored non-interactive methods for acquiring POI attributes from web pages~\cite{sun2021gedit}, street views~\cite{fan2022dumapper}, and user-generated content~\cite{welty2022addressing}. However, these non-interactive methods suffer from several challenges: (i) The data sources may cover only limited POIs and are prone to becoming outdated. (ii) Extracting POI information from open data sources is challenging due to the free-form nature of the textual and visual data. (iii) These approaches typically require high labor costs in the post-prepocessing phase, hindering rapid system updates. 

To tackle these challenges, the DuIVRS-1 system has been proposed, a telephonic Interactive Voice Response (IVR) solution for POI attribute acquisition~\cite{huang2022duivrs}. It proactively gathers POI attributes by automatically initiating calls and engaging in conversations with POI owners to collect information. As shown in Figure~\ref{duivrs}, DuIVRS-1 employs a sequential modular design, including natural language understanding (NLU)~\cite{namazifar2021language}, dialogue management (DM)~\cite{carfora2020dialogue}, and natural language generation (NLG)~\cite{peng2020few}, with each module focusing on a specific subtask. 
However, the modular design is prone to error accumulation and complicates system maintenance.

Recently, LLMs have achieved impressive performance across various applications~\cite{zhang2026omegause,wang2026aradd}. 
However, the industrialized applications of LLM on IVR systems are still limited due to LLM's higher cost, slower inference speed, and difficulty in ensuring stability. Based on this, we focus on how to implement industrialized application deployment of LLM on IVR systems for POI attribute acquisition. 
Specifically, we introduce DuIVRS-2, a reshaped version of the original DuIVRS system, incorporating an LLM to enable a transition towards an end-to-end architecture. 
To begin, we propose a data augmentation module designed to synthesize the conversation between users and assistants, thereby enhancing data diversity and mitigating issues related to data imbalance.
By fine-tuning an LLM on these data, we enable the LLM to adapt effectively to our specific scenarios. 
Additionally, we streamline the conversation generation process through a selective generation scheme and incorporate a CoT mechanism. These enhancements improve the stability and controllability of LLM's output.
Furthermore, we design a cooperative iterative learning strategy to progressively enhance the model's performance by alternating data growth and policy improvement.
To facilitate deployment, 
we implemented an LLM acceleration strategy to efficiently manage time delays, enhancing the system's responsiveness.
Our comprehensive evaluations, conducted both offline and online, demonstrate that DuIVRS-2 significantly outperforms its predecessor.

\section{Related Work}

\subsection{POI Attribute Acquisition}

The acquisition of POI attributes is crucial for various downstream tasks, including POI retrieval~\cite{yuan2020spatio, huang2021hgamn} and POI recommendation~\cite{zhao2020go}. 
In this domain, existing research has identified four primary sources for extracting POI information:
(i)  \textbf{Street Views}: Primarily utilized to gather the names and locations of POIs. However, this method is limited in obtaining more detailed information~\cite{revaud2019did, fan2022dumapper}.
(ii)  \textbf{Web Pages}: These sources acquire POI attributes using natural language processing techniques to parse unstructured web pages~\cite{xu2019dlocrl, sun2021gedit, yang2022point}. Nonetheless, their coverage is restricted, with less than 30\% of POIs having an associated website.
(iii) \textbf{User-Generated Content}: An example of this method is Google's Local Guide System in Maps~\cite{welty2022addressing}. Although reliant on user contributions, this approach covers less than 10\% of POIs and is often hindered by issues of timeliness.
(iv)\textbf{IVRS-Based Methods}: This strategy obtains detailed POI attributes by engaging POI owners through telephonic interactive voice response systems. DuIVRS, an entirely automated solution featuring machine-directed dialogues, has been successfully implemented in Baidu Maps, showing superior performance~\cite{huang2022duivrs}.
Given the impossibility of capturing all attributes for all POIs, these methods serve complementary roles in practice. In this paper, we build upon IVRS-based techniques to introduce DuIVRS-2, an advanced iteration of the original DuIVRS system. This version integrates an LLM to facilitate a shift from a traditional modular design to an end-to-end architecture.

% \vspace{-0.2cm}
\subsection{Task-oriented Dialogue System}

Task-Oriented Dialogue (TOD) systems are designed to achieve specific objectives within a particular domain~\cite{ni2023recent}. 
Traditional TOD systems generally follow a pipeline with four components: Natural Language Understanding (NLU) to categorize user intentions, Dialogue State Tracking (DST) to maintain the state of the dialogue, Policy Learning (POL) to determine the agent's next move, and Natural Language Generation (NLG) to translate actions into language. 
Recent research focuses on streamlining the modeling process through neural network models adept at handling TOD sub-tasks, including text generation~\cite{budzianowski2019hello, wang2022task} and framing all sub-tasks as sequence prediction~\cite{hosseini2020simple}. 
The rise of LLMs has significantly shifted the development of TOD systems~\cite{sun2021ernie, achiam2023gpt}, moving towards LLM-based end-to-end dialogue systems instead of traditional modular architectures. 
Relevant work includes using LLM prompts for managing multi-participant dialogues~\cite{mao2024multi} and developing privacy-preserving chatbots for chronic disease support~\cite{montagna2023data}. 
Despite these efforts, in practical deployments, existing methods are typically constrained by industrial production limits, stability, and latency requirements. Thus, in DuIVRS-2, we innovatively incorporate an LLM into the interactive voice response system to meet the demands for rapid online deployment response while achieving high precision in POI attribute retrieval.

\section{Preliminary}

\subsection{Essential Terminologies}

To frame the subsequent discussion, we categorize LLMs into three types: \textbf{LLM-S}mall, \textbf{LLM-L}arge, and \textbf{Black-box LLM}.
Specifically, LLM-S denotes an LLM with fewer parameters (typically less than 2 billion), while LLM-L denotes an LLM with a significantly larger number of parameters. 
Despite differing in size, both LLM-S and LLM-L follow a similar architectural framework and can be fine-tuned for specific tasks. Generally, LLM-L surpasses LLM-S in performance metrics, albeit at a lower inference speed.
Black-box LLM represents exceptionally large and proprietary models like GPT-5 and \eb. These models exhibit broad capabilities, enabling high performance across a diverse range of tasks.

\begin{figure*}[t]
\centering
\includegraphics[width=0.85\linewidth]{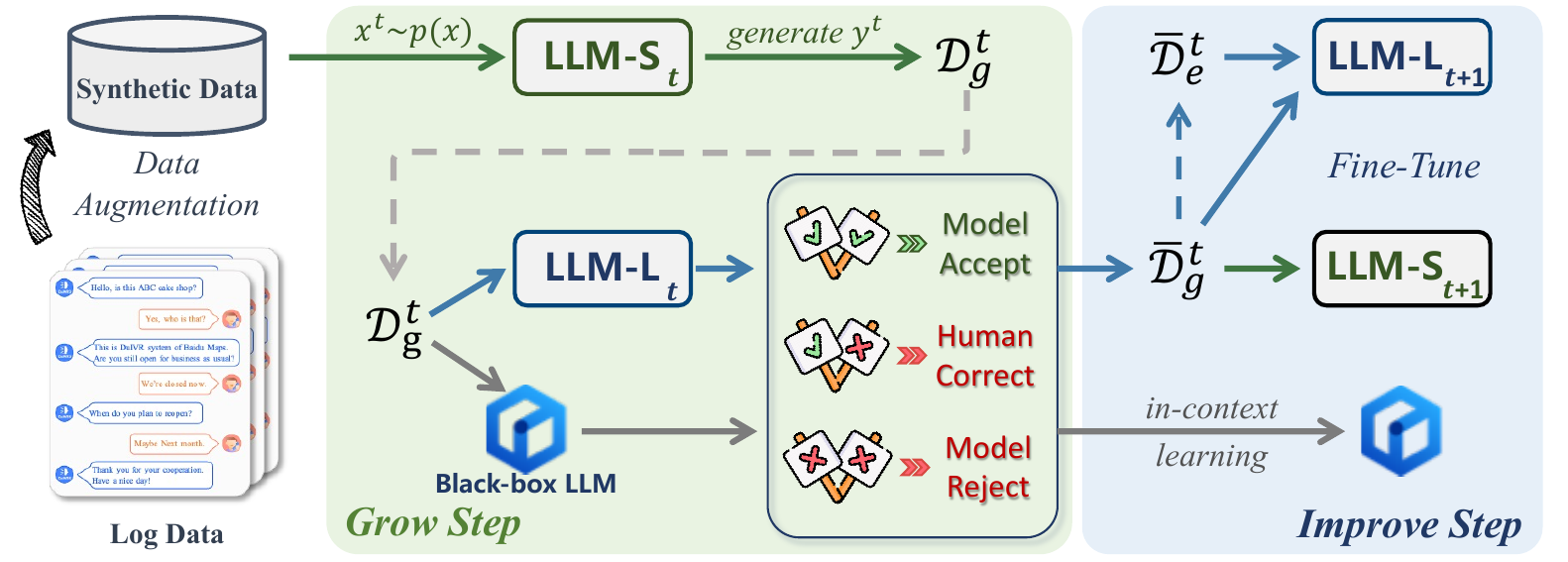}
\caption{The cooperative iterative learning in DuIVRS-2,  which operates via an iterative two-step process: (1) the grow step augments data and generates responses evaluated by LLM-L and a Black-box LLM, with uncertain cases reviewed by humans; (2) the improve step updates LLM-S and LLM-L using accepted responses, while also refining the Black-box evaluator through in-context learning.
}
\label{duivrs-2}
\end{figure*}

\subsection{Problem Formulation}
We focus on an industrial IVR application for POI attribute acquisition. The system acts as an agent initiating a call to a POI owner to verify or acquire specific attributes (e.g., business hours, location).

Let $C_{[1,m]} = (u_1, s_1, \dots, u_m, s_m)$ denote a truncated dialogue history, where $u_i$ is the agent-issued query and $s_i$ is the POI owner's response at turn $i$. The objective is to generate the next agent query $u_{m+1}$ that maximizes the information gain regarding the target POI attributes, while maintaining conversational naturalness.

Unlike standard open-domain generation, this task is constrained by:

\begin{enumerate}
    \item \textbf{Safety:} The agent must not hallucinate non-existent attributes.
    \item \textbf{Latency:} The response $s_{m+1}$ must be generated within strict time limits (typically < 200ms) to prevent user hang-ups.
\end{enumerate}
\section{DuIVRS-2 System}

This study focuses on the development of an end-to-end LLM-based agent tailored for acquiring POI attributes through telephonic
conversations.

\subsection{Addressing Data Imbalance via FSM-Guided Augmentation}
\label{subsection:DA}

\begin{figure}[t!]
    \centering
    \begin{subfigure}{0.45\textwidth}
        \centering
        \includegraphics[width=\textwidth]{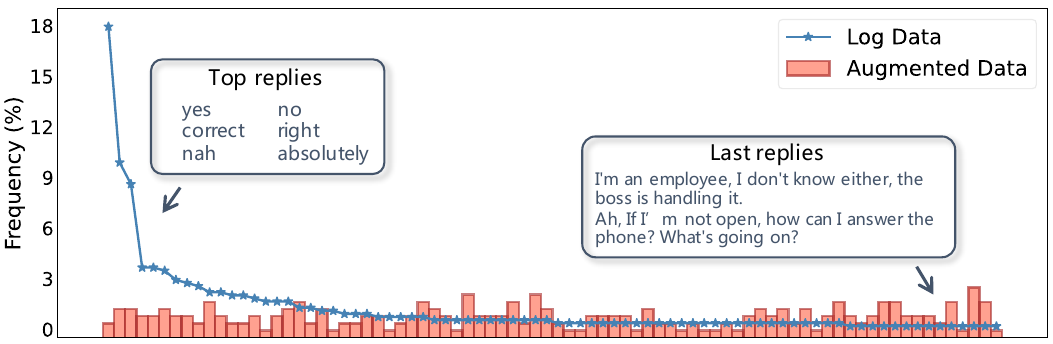}
        \caption{Distribution of user replies.}
    \end{subfigure}\hfill
    \begin{subfigure}{0.45\textwidth}
        \centering
        \includegraphics[width=\textwidth]{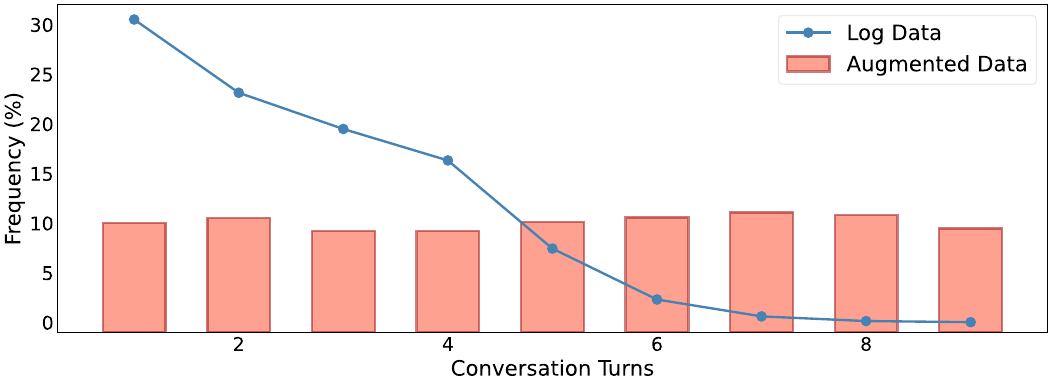}
        \caption{Distribution of conversation turns.}
    \end{subfigure}
    \vspace{-3mm}
    \caption{Comparison of distributions before (log data) and after FSM-guided augmentation, highlighting the mitigation of long-tail issues.}
    \vspace{-3mm}
    \label{fig:frequency}
\end{figure}

A primary challenge in training dialogue systems on historical production logs is the data distribution. As shown in Figure~\ref{fig:frequency}, user replies and conversation turns in our legacy system (DuIVRS-1) logs exhibit a pronounced long-tail distribution. Direct fine-tuning on this data causes the model to overfit to frequent, simple interactions while failing on critical edge cases.

To construct a balanced training set, we devised a data augmentation strategy using a Finite State Machine (FSM) derived from the legacy rule-based system. We mapped the legacy system's fixed response templates to FSM states ($S$) and user replies to transitions ($\Sigma$). By parsing years of historical logs, we extracted the empirical transition patterns driven by real user behavior. More details are shown in Appendix ~\ref{ill_fsm}

We re-balanced the data distribution by sampling from this FSM structure. Instead of sampling based on raw log frequency, we enforced a uniform distribution strategy:
\begin{enumerate}
    \item \textbf{Path Sampling:} We extract all valid state transition sequences (dialogue paths) and group them by length. We uniformly sample across length categories to ensure the model learns both short and long-range dependencies.
    \item \textbf{Transition Sampling:} Between states, we uniformly sample user replies from the set of historical variations, ensuring diverse lexical coverage.
\end{enumerate}

As illustrated in Figure~\ref{fig:frequency}, this approach shifts the training data from a long-tailed to a uniform distribution, effectively solving the "cold start" problem for rare dialogue states without requiring expensive new data collection.

\subsection{Latency-Efficient LLM-Based Dialogue Management}

In a telephonic setting, inference latency is a hard constraint. Consequently, we utilize a smaller, efficient LLM (denoted as \textbf{LLM-S}) rather than a massive general-purpose model. To ensure safety and tractability given the smaller parameter size, we employ a selective generation approach.

Rather than allowing open-ended generation, which poses safety risks (hallucinations), we structure the input prompt $\textbf{X}$ to include the conversation history and a set of valid \textit{Reply Options} derived from the FSM transitions. The output $\textbf{Y}$ is formulated as a CoT reasoning step followed by the selection of the most appropriate option. This formulation allows the model to explicitly classify user intent before selection, significantly improving interpretability and robustness.

\subsection{Cooperative Iterative Learning}

Historical logs often contain ASR errors or misinterpretations by the legacy system. Training on this noisy data limits performance. We propose a Human-in-the-Loop cooperative iterative learning framework to clean the data and refine the model, inspired by the ReST approach~\cite{gulcehre2023reinforced}, which is shown in Figure~\ref{duivrs-2}.

\subsubsection{Dual-LLM Cooperative Evaluation}
To scale data quality without prohibitive manual annotation costs, we construct an automated evaluation ensemble consisting of two distinct evaluators:

\textbf{LLM-L for Evaluation.} Given the same training data and task format, LLM-L is expected to have stronger comprehension and generation capability than LLM-S due to its larger capacity~\cite{touvron2023llama}. We therefore fine-tune LLM-L on the same domain-specific data and use it to evaluate LLM-S's outputs from both generative and discriminative perspectives.

From the generative perspective, we evaluate the LLM-S-generated output $\mathbf{Y}$ by computing its conditional likelihood under LLM-L given the input $\mathbf{X}$. The normalized sequence likelihood is:
\begin{equation}
\small
    P_{\mathrm{gen}}(\mathbf{Y}|\mathbf{X})
    =
    \prod_{i=1}^{|\mathbf{Y}|}
    p_{\mathrm{LLM\text{-}L}}(\mathbf{Y}_i|[\mathbf{X}, \mathbf{Y}_{1:i-1}])^{\frac{1}{|\mathbf{Y}|}},
\end{equation}
where $\mathbf{Y}_i$ denotes the $i$-th token of $\mathbf{Y}$, and $p_{\mathrm{LLM\text{-}L}}(\cdot)$ is the next-token probability predicted by LLM-L.

From the discriminative perspective, we further train LLM-L to judge whether an input-output pair is correct. For each positive pair $(\mathbf{X}, \mathbf{Y})$, we construct a negative pair by replacing $\mathbf{Y}$ with an incorrect output $\hat{\mathbf{Y}}$. The evaluation dataset is formulated as follows:
\begin{equation}
    \mathcal{D}_e^t
    =
    \{((\mathbf{X}, \mathbf{Y}), 1), ((\mathbf{X}, \hat{\mathbf{Y}}), 0)\}.
\end{equation}
We then format each pair into an evaluation prompt and fine-tune LLM-L to output the corresponding binary label. The discriminative confidence is defined as follows:
\begin{equation}
\small
    P_{\mathrm{disc}}(1|[\mathbf{X}, \mathbf{Y}^*])
    =
    \frac{\exp(\bar{g}(1|\mathbf{X}_e))}
    {\sum_{l \in \{0,1\}}\exp(\bar{g}(l|\mathbf{X}_e))},
\end{equation}
where $\mathbf{X}_e$ denotes the evaluation prompt, and $\bar{g}(l|\mathbf{X}_e)$ is the logit assigned to label $l$.

Finally, we combine the generative and discriminative scores to obtain the confidence score assigned by LLM-L:
\begin{equation}
\label{LLM-L-judge}
    c
    =
    (1-\alpha) \cdot P_{\mathrm{gen}}(\mathbf{Y}|\mathbf{X})
    +
    \alpha \cdot P_{\mathrm{disc}}(1|[\mathbf{X}, \mathbf{Y}]),
\end{equation}
where $\alpha$ is a hyperparameter balancing the two evaluation perspectives.

\textbf{Black-box LLM for Evaluation.} To prevent "inbreeding" where LLM-L and LLM-S reinforce shared errors (e.g., specific ASR noises present in the training data), we utilize an independent Black-box LLM (e.g., ERNIE 4.0) as an unbiased judge. This model evaluates samples via in-context learning with detailed reasoning prompts.

\subsubsection{The Iterative Loop}
The training process alternates between a \textit{Grow Step} and an \textit{Improve Step}:
\begin{itemize}
    \item \textbf{Grow:} We sample new dialogues using the current policy $\pi_t$. The dual-evaluator (LLM-L + Black-box) filters these samples. High-confidence samples are added to the training set automatically. Samples with improved scores or disagreement between evaluators are routed to human annotators for correction.
    \item \textbf{Improve:} The model $\pi_{t+1}$ is fine-tuned on this cleaned, augmented dataset ($\bar{\mathcal{D}}_g^t$). Simultaneously, the feedback from human annotators is used to update the prompts for the Black-box evaluator and the fine-tuning of LLM-L.
\end{itemize}

This cycle creates a "data flywheel", progressively removing noise from the historical logs and adapting the model to complex real-world scene.

\section{Experiments}

\subsection{Experimental Setup}

\noindent \textbf{Datasets.}
For offline training, we adopted the data augmentation strategy in Section~\ref{subsection:DA}. The initial training set included 5,000 dialogues, with 5,000 additional samples per iteration during fine-tuning. We evaluated the model using three test sets derived from DuIVRS-1 logs: $D_{\text{effect}}$, $D_{\text{general}}$, and $D_{\text{robust}}$.

\noindent \textbf{Evaluation Metrics.}
\label{tsr} We employed two primary metrics: (1) Consistency Rate (CR) to evaluate offline single-turn effectiveness, and (2) Task Success Rate (TSR) to assess online multi-turn interaction performance.

\noindent \textbf{Experimental Setting.}
We chose ERNIE-Bot-tiny (EB-tiny) to serve as our LLM-S, ERNIE-Bot-turbo (EB-turbo) as our LLM-L, and ERNIE 4.0 as our Black-box LLM. More experimental details are provided in Appendix~\ref{appendix:exp_details}.

\subsection{Offline Evaluation}
\label{sec:overall_perform}

\begin{table}[t]
\centering
\small
\caption{Offline performance on CR metric.}
\vspace{-0.2cm}
\setlength{\tabcolsep}{4pt}
\begin{tabular}{c|cccc}\toprule
\textbf{Model} & \textbf{$D_{\text{effect}}$} & \textbf{$D_{\text{general}}$} & \textbf{$D_{\text{robust}}$} & \textbf{Avg.} \\ \midrule
DuIVRS-1  & 72.20\% & 62.99\% & 69.05\% & 68.08\% \\
GPT-4o & 72.11\%  &  64.81\%  &  63.13\% &  66.68\% \\
DeepSeek-V3 & 73.18\% & 63.87\% & 64.55\% &  67.20\% \\
HybridLLMs & 81.37\% & 73.83\% & 75.89\% & 77.03\% \\
\textbf{DuIVRS-2}  & \textbf{81.62}\% & \textbf{73.70}\% &\textbf{76.22}\% & \textbf{77.18}\% \\ \midrule
\textit{Ablations} & & & & \\
% \multicolumn{5}{l}{\textit{Ablation Studies}} \\
LLM-DM & 76.68\% & 67.40\% & 60.98\% & 68.35\% \\
Direct-SFT & 69.18\% & 60.33\% & 52.90\% & 60.80\% \\
w/o-CoT  & 44.25\% & 44.41\% & 28.35\% & 39.00\% \\
w/o-DA & 76.52\% & 62.36\% & 54.12\% & 64.33\% \\ \bottomrule
\end{tabular}
\label{offline}
\vspace{-4mm}
\end{table}

We compared DuIVRS-2 with its predecessor, DuIVRS-1, which has been deployed in Baidu Maps for several years and has proven its effectiveness. Additionally, we included GPT-4o and DeepSeek-V3~\cite{liu2024deepseek}, two of the most advanced general-purpose LLMs, as strong baselines. Beyond these, we have not identified other task-oriented dialogue systems that can be directly deployed for our task.

As shown in Table~\ref{offline}, DuIVRS-2 demonstrates substantial improvements across all evaluation sets, achieving an average performance gain of 13.37\% over DuIVRS-1, 14.85\% over DeepSeek-V3, and 15.74\% over GPT-4o. 
These results validate the effectiveness of our task-specific dialogue management framework in diverse interaction scenarios. 
In contrast, GPT-4o and DeepSeek-V3, while capable of general reasoning, show limited performance on long-tail and domain-specific POI tasks due to a lack of adaptation.

To further evaluate the generalizability and robustness of our framework across different model families, we introduce an additional baseline named \textbf{HybridLLMs}. In this configuration, LLM-S and LLM-L are implemented using Qwen2.5-1.5B and Qwen2.5-7B~\cite{team2024qwen2}, respectively, and the Black-box LLM is replaced with GPT-4o. Notably, HybridLLMs achieves an average CR of 77.03\%, which is within 0.15 percentage points of DuIVRS-2, and exhibits consistent performance across all test sets. This empirical result demonstrates that our framework is largely model-agnostic, attributing the performance gains of DuIVRS-2 to its architectural design, including FSM-guided data augmentation, selective generation with CoT, and cooperative iterative learning, rather than any specific model series.
Besides, this finding helps address concerns regarding potential \textit{model-series bias}, as all three models in DuIVRS-2 (LLM-S, LLM-L, and Black-Box LLM) originate from the ERNIE family. While that design choice was driven by internal deployment efficiency and ecosystem compatibility, the HybridLLMs experiment confirms that our framework maintains comparable performance when applied to heterogeneous models.

\subsubsection{Non-Iterative Ablation Study}
In this part, we evaluate our LLM-based Dialogue Management (LLM-DM) against its variants prior to the cooperative iterative learning framework. Variants include:

\textbullet~\textbf{Direct-SFT:} Directly generating next-round query sentence.

\textbullet~\textbf{w/o-DA:} LLM-DM excluding data augmentation techniques.

\textbullet~\textbf{w/o-CoT:} LLM-DM generating the next response choice without the reasoning process.

The results, detailed in Table~\ref{offline}, reveal significant insights. Compared with DuIVR-1, our basic model, LLM-DM, fine-tuned on a limited dataset, even without the iterative refinement, can achieve comparable performance, especially in terms of $D_{\text{effect}}$ and $D_{\text{general}}$ datasets. This supports our hypothesis that LLM has the potential to transform the DuIVRS-1 into an end-to-end architecture.
When compared with its variants, LLM-DM, incorporating all elements, outperforms the others, highlighting the effectiveness of our comprehensive system design.
Direct-SFT exhibits poor performance, indicating the necessity of a tailored training scheme for LLMs.
The w/o-DA variant shows a slight decline in performance, particularly in $D_{\text{general}}$ and $D_{\text{robust}}$ datasets, underscoring the value of the data augmentation module in enhancing model performance in uncommon scenarios.
The w/o-CoT variant performs the poorest, illustrating the critical role of the reasoning process.

\subsubsection{Iterative Learning Analysis} Here, we focus on evaluating the cooperative iterative learning strategy of DuIVRS-2 and its variants:

\textbullet~\textbf{w/o-\eb:} It is a variant which DuIVRS-2 removes the Black-box LLM evaluator.

\textbullet~\textbf{w/o-EB-turbo:} Here, DuIVRS-2 excludes the LLM-L evaluator.

\begin{figure*}[t!]
  \centering
  \begin{subfigure}[b]{0.3\textwidth}
    \includegraphics[width=\textwidth]{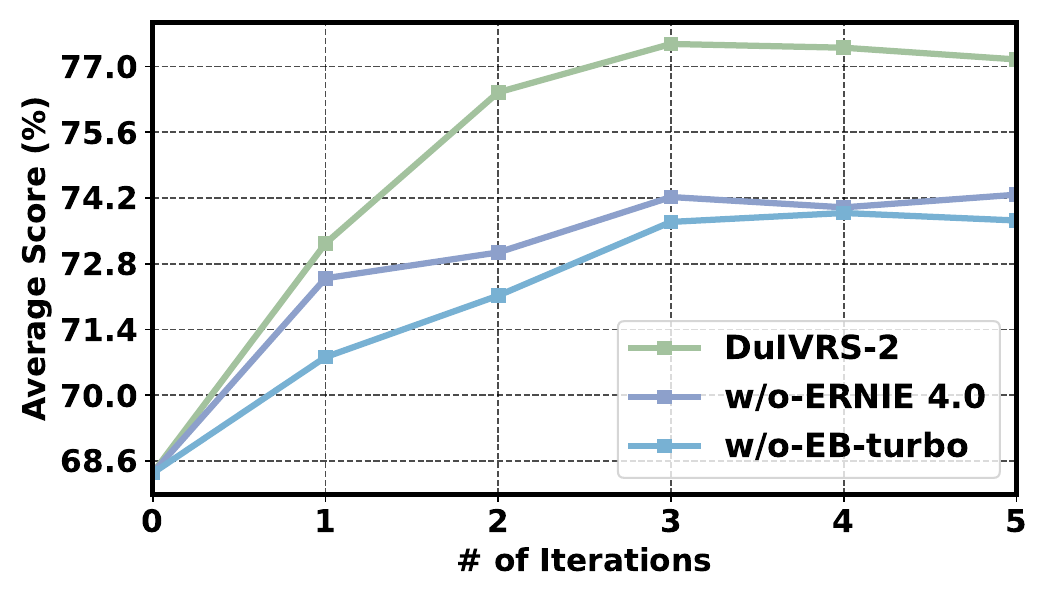}
    \caption{Average Score.}
    \label{fig:iteration_performance}
  \end{subfigure}
  \hfill
  \begin{subfigure}[b]{0.3\textwidth}
    \includegraphics[width=\textwidth]{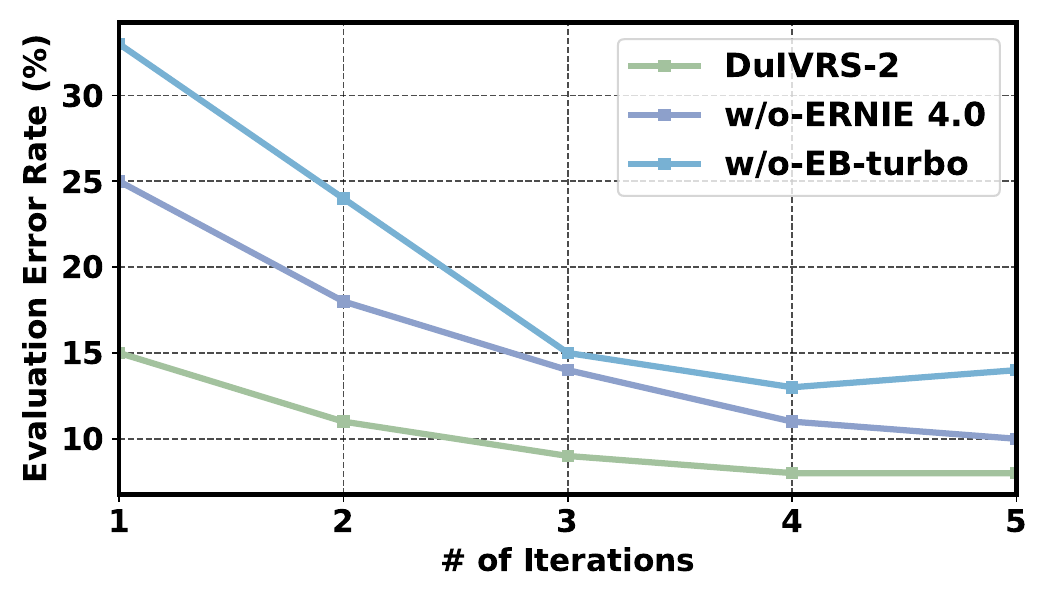}

    \caption{Evaluation Error Rate.}
    \label{fig:evaluation_error_rate}
  \end{subfigure}
  \hfill
  \begin{subfigure}[b]{0.3\textwidth}
    \includegraphics[width=\textwidth]{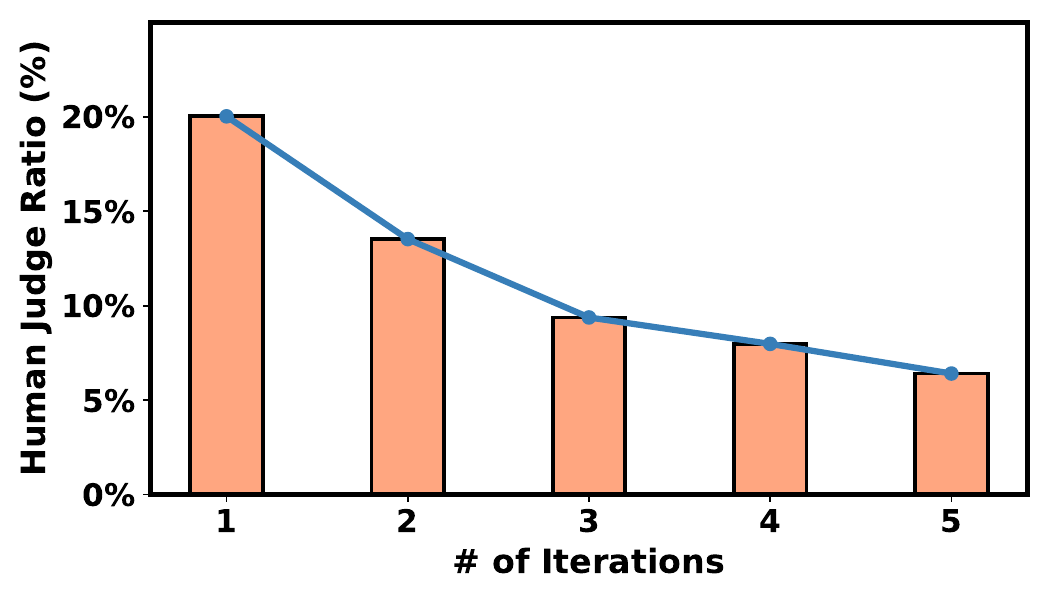}

    \caption{Human Judge Ratio.}
    \label{fig:human_judge_ratio}
  \end{subfigure}
  \vspace{-4mm}
  \caption{Main properties change with the iteration process.}
  \vspace{-4mm}
  \label{fig:iteration}
\end{figure*}

As shown in Figure~\ref{fig:iteration_performance}, the observed trend indicates that as the number of iterations increases, the model's performance gradually improves, which emphasizes the significance iterative process. After a few iterations, the performance gains began to plateau, reaching a point of stability. This pattern suggests that in practical applications, it is efficient to iterate the model 3-4 times to achieve optimal performance enhancements. Moreover, the removal of either evaluator results in a noticeable performance decline throughout the iteration process, which shows the effectiveness of both evaluators for the cooperative reward scheme.

We further evaluate the effectiveness of DuIVRS-2's evaluation module by measuring the evaluation error rates across different system variants and training iterations, as shown in Figure~\ref{fig:evaluation_error_rate}. The results indicate a steady decline in error rates over successive iterations, reflecting improved evaluation accuracy through iterative refinement. Among the single-evaluator variants, w/o-ERNIE 4.0 (i.e., using LLM-L as the sole evaluator) slightly outperforms w/o-EB-turbo (i.e., using the Black-box LLM alone), highlighting the advantage of domain-specific fine-tuning. Notably, the full model that combines both evaluators via a cooperative voting mechanism achieves the lowest error rate, demonstrating the effectiveness and robustness of the dual-evaluator strategy.

Regarding labor costs, as shown in Figure~\ref{fig:human_judge_ratio}, the proportion of data requiring human judgment steadily declines across iterations. This reduction is primarily enabled by the cooperative reward mechanism, which allows the majority of samples to be automatically evaluated, thereby substantially reducing reliance on human intervention throughout the iterative learning process.

\subsection{Deployment Preparation}
Following the successful offline validation of DuIVRS-2’s efficacy, we proceeded to its online deployment phase. For this crucial step, we selected the LLM-S model trained through the cooperative iterative learning framework, owing to its compact architecture, controllable behavior, and low inference overhead. Specifically, the LLM-S model contains fewer than 2 billion parameters and supports selective generation with minimal latency.

To ensure deployment readiness, we performed several system-level optimizations. First, the dynamically constructed computation graph was converted into a static format using the PaddlePaddle framework\footnote{https://www.paddlepaddle.org.cn/}, which significantly reduced runtime overhead by eliminating graph recompilation. Additionally, we applied int8 quantization to the model weights, achieving substantial memory savings and further improving inference throughput.

Moreover, we integrated FastDeploy\footnote{https://github.com/PaddlePaddle/FastDeploy}, an inference optimization platform, to streamline runtime performance. FastDeploy allows efficient resource scheduling and batch management, which is crucial for high-frequency, large-scale call services. As in DuIVRS-1~\cite{huang2022duivrs}, we retained the proven ASR and TTS modules to maintain robustness in audio-based interaction.

\subsection{Online Evaluation}
\label{online_performance}

To evaluate real-world applicability, we conducted a two-month online A/B test involving three systems: human operators, DuIVRS-1, and DuIVRS-2. Incoming calls were randomly partitioned among these systems, with DuIVRS-2 allocated a fixed quota of approximately 3,000 calls per day during a controlled one-hour window. This conservative traffic setting allowed us to validate system stability and assess performance under moderate load, while minimizing potential disruption to user experience and business operations.

We collected and analyzed logs from the A/B test period to compare DuIVRS-2 with both its predecessor (DuIVRS-1) and human operators. As shown in Table~\ref{online}, DuIVRS-2 achieved a TSR (see Subsection~\ref{tsr}) of 83.9\%, outperforming DuIVRS-1 by 4 percentage points and reaching 93.64\% of human performance. Despite incorporating an LLM, DuIVRS-2 maintained a cost per call below ¥0.2—comparable to DuIVRS-1 and significantly lower than human-operated calls (~¥1.5).
Importantly, DuIVRS-2 achieved an average reaction time of 130ms, which remains well within acceptable latency bounds for real-time dialogue systems (typically <200ms for human perception). This confirms its suitability for responsive interaction in production environments. Moreover, DuIVRS-2 can handle up to 0.4 million calls per day, vastly surpassing the throughput limit of human agents ($\leq$ 200 calls/day), and supporting truly large-scale deployment scenarios.

Together, these results confirm that DuIVRS-2 meets the stringent latency, scalability, and cost requirements necessary for industrial-grade, large-scale POI attribute acquisition.

\begin{table}[t]
\centering
\small
\caption{Comparision between human and DuIVRSs.}
\begin{tabular}[t]{c|ccc}\toprule
\textbf{metric} & \textbf{Human} & \textbf{DuIVRS-1} & \textbf{DuIVRS-2} \\ \midrule
TSR & 89.6\% & 79.9\% & 83.9\% \\
cost per call & \textyen 1.5 &  <\textyen 0.2 & <\textyen 0.2 \\
reaction time & 500ms & 15ms & 130ms \\
call per day & $\leq$ 200 & no limitation & 0.4 million \\
scalability  &  \ding{55} & \ding{51} & \ding{51} \\
\bottomrule
\end{tabular}
% \end{adjustbox}
\label{online}
\end{table}

\section{Conclusion}

In this paper, we proposed DuIVRS-2, an advanced system for POI attribute acquisition, which leverages LLM to transform the traditional modular IVR system into an end-to-end architecture. Specifically, we developed an innovative LLM-based framework, meticulously designed to ensure stability and controllability in the output. And the framework is optimized for low latency, enabling real-time responses, and incorporates mechanisms for continuous refinement. Our comprehensive evaluations demonstrate that DuIVRS-2 significantly outperforms its predecessor. By deploying it in Baidu Maps for two months, we observed that DuIVRS-2 surpassed DuIVRS-1 by 4 percentage points in task success rate, accompanied by a minor increase in operational costs.

\section*{Acknowledgments}
This research was supported in part by the National Natural Science Foundation of China under Grant No.92370204.

\bibliography{main}

\appendix
\section{Appendix}\nopagebreak

\subsection{Experimental Details} \label{appendix:exp_details}

\subsubsection{Detailed Dataset Sampling Strategies}
To comprehensively evaluate the performance of DuIVRS-2, we constructed three test datasets from the original DuIVRS-1 logs using the following sampling strategies:
\begin{itemize}
    \item \textbf{$D_{\text{effect}}$}: Samples are drawn according to the natural distribution of user replies in the production logs. This dataset evaluates the performance in typical, everyday user interactions.
    \item \textbf{$D_{\text{general}}$}: Each unique user reply is sampled with equal probability, regardless of its frequency in the logs. This ensures that less frequent but critical scenarios are adequately represented to test generalization.
    \item \textbf{$D_{\text{robust}}$}: Specifically selects user replies that are characterized by longer length or semantic complexity. This evaluates the model's ability to handle noisy inputs and complex intentions.
\end{itemize}

\subsubsection{Mathematical Definitions of Evaluation Metrics}
The performance is quantified using the following formal definitions:

\begin{itemize}
    \item \textbf{Consistency Rate (CR)}: Calculated for single-turn dialogues as:
    $$CR = \frac{N_C}{N} \times 100\%$$
    where $N$ is the total number of evaluated dialogues, and $N_C$ is the number of responses judged correct by human annotators based on reasoning alignment and query consistency.
    
    \item \textbf{Task Success Rate (TSR)}: Calculated for multi-turn interactions as:
    $$TSR = \frac{N_S}{N_T} \times 100\%$$
    where $N_T$ is the total number of query prompts issued by the system during a full dialogue session, and $N_S$ is the number of successful attribute acquisitions (e.g., confirming a POI).
\end{itemize}

\subsubsection{Training Details}
The model training was conducted on Baidu's PaddleCloud, utilizing eight NVIDIA A100-80G GPUs for fine-tuning.
We utilized the AdamW~\cite{loshchilov2017decoupled} optimizer with the parameters set to ${\beta}_1=0.9, {\beta}_2=0.95, eps=1e-5$. 
The training process was configured with a batch size of 128 and a sequence length limited to 1024 tokens.
A linear learning rate schedule was applied, incorporating a warm-up phase covering 3\% of the total training steps. 
The maximum learning rates were established at $2 \times 10^{-5}$ for the EB-turbo model and $1 \times 10^{-4}$ for the EB-tiny model.
For the EB-tiny model, we employed bf16 16-bit (mixed) precision for full-parameter fine-tuning. For the EB-turbo model, we use Lora~\cite{hu2021lora} for parameter-efficient fine-tuning.
Finally, we fine-tuned the two models for 2 epochs.

\subsection{Additional Analysis}
\subsubsection{Latency and Efficiency Analysis}
\label{sec:latency}
During the online deployment, DuIVRS-2 is hosted on eight NVIDIA A10 GPUs (24GB memory per GPU) using the Triton Inference Server. With 8-bit quantization applied to the LLM-S model, the system achieves a memory footprint of approximately 22GB per GPU. Under this configuration, the model attains an average inference latency of about 130ms  per query and a throughput of 61.5 queries per second (QPS) per GPU. 

In addition, an important factor contributing to latency stability is the system’s outbound calling architecture. Since call concurrency is scheduled and initiated from the backend, DuIVRS-2 avoids unpredictable user-driven load surges that are common in inbound call centers. This allows the system to maintain consistent response times and system utilization, even during peak operating periods.

\subsubsection{Stability Analysis}
\label{stability_analysis}

To evaluate the impact of selective generation and reasoning on output stability, we conduct ablation studies comparing DuIVRS-2 with two simplified variants: Direct-SFT and w/o-CoT. In our setting, we define a hallucination as any AI-generated query that either contradicts the dialogue context or falls outside the FSM-defined set of candidate responses at a given turn, thereby capturing both semantic inconsistency and structural deviation from valid system behavior.

To quantify hallucination rates, we perform human evaluation on sampled interactions. For Direct-SFT, annotators flag a response as hallucinated if it introduces a query that is logically disconnected from prior user inputs or invents details absent from the dialogue history. For DuIVRS-2 and w/o-CoT, hallucinations are identified when the model’s output is not among the FSM-permitted candidate options at a specific turn. We illustrate representative hallucination cases in Appendix~\ref{sec:hallucination}, highlighting distinct failure modes.

Our findings show hallucination rates of 0\%, 1.30\%, and 2.08\% for DuIVRS-2, Direct-SFT, and w/o-CoT, respectively. These results demonstrate the effectiveness of DuIVRS-2’s selective generation strategy in maintaining semantic grounding. Notably, the w/o-CoT variant exhibits the highest hallucination rate, underscoring the critical role of CoT-style reasoning in ensuring output stability. 

In real-world deployment, DuIVRS-2 also employs a fallback mechanism to guard against hallucinated outputs. When a generated response falls outside the FSM-defined candidate set, the system repeats the last valid response up to three times. If invalid output persists, it reissues the original question instead—preserving conversational coherence and preventing logical drift.

\subsubsection{Power Consumption.}
Building on prior research~\cite{wu2022sustainable, touvron2023llama2} and power consumption data for GPU devices, we aim to estimate the financial costs and carbon emissions associated with our training process. 
Along with previous work, our analysis excludes additional power requirements, such as those from interconnects or ancillary non-GPU energy expenditures. 
At each iterative stage, the training duration for EB-tiny is about 1 hour, and EB-turbo is 14 hours, amounting to a cumulative $(1+14) \times 8 \times 6 = 720$ GPU hours on A100-80G units with a TDP of 400W. Considering GPUs' actual power use (typically under 400W) and an electricity rate of 1.2 RMB/kWh, the maximum training expense is roughly 400 RMB, with carbon emissions approximating 122kgCO2eq. 
Furthermore, utilizing the Ernie 4.0 API service adds to the cost. With a rate of 0.15 RMB per 1k tokens and across five iterations totaling 72,000 requests at 0.5k tokens each, the API costs about $0.15 \times 72 \times 10^3 = 5400$ RMB. 
Overall, the complete training costs are projected to stay below 10,000 RMB, with carbon emissions under 1tkgCO2eq.

\begin{figure}[t!]
\centering
\includegraphics[width=0.7\linewidth]{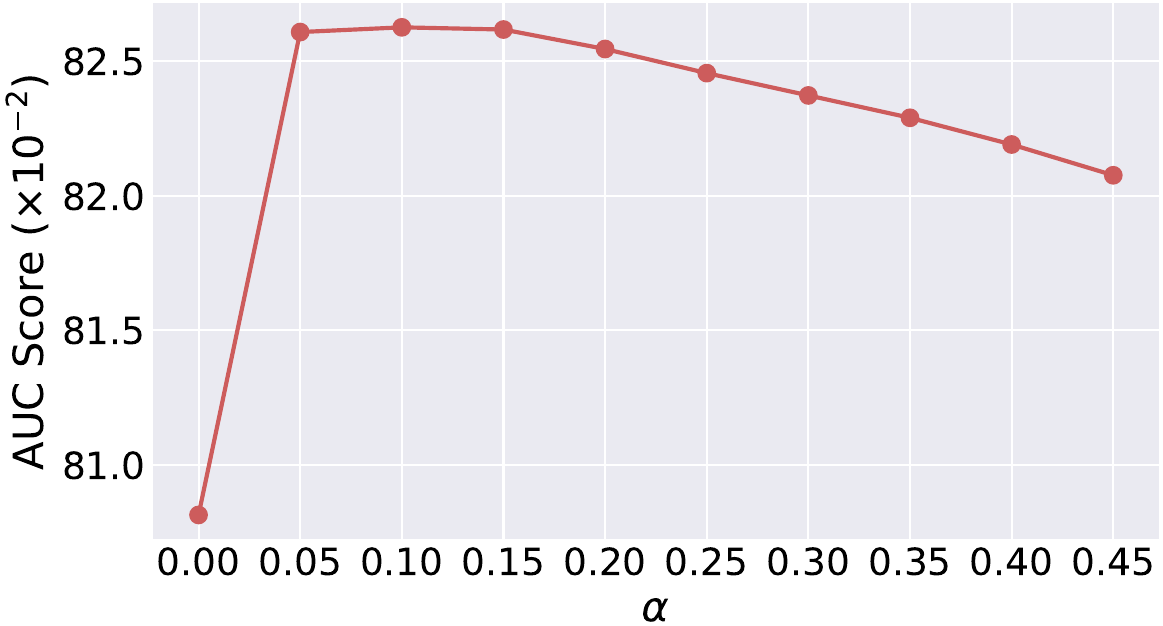} 
\caption{Parameter sensitivity analysis of $\alpha$.}
\label{fig:parameter_sensitivity}
\end{figure}

\subsection{Case Study}
We present a case study to illustrate the superiority of DuIVRS-2 in navigating certain edge cases, compared to its predecessor DuIVRS-1 and \eb.
The example extracted from online test showcases a scenario where a user responds with a counter-question to an AI inquiry about the business status, rather than providing a direct response. While DuIVRS-1 misinterprets this rhetorical response as an indication of the business being closed and asks for the reopening time. \eb recognizes the rhetorical nature but still fails to grasp the underlying intent of the user, leading them to seek confirmation about the business status. 
DuIVRS-2, leveraging its rich domain knowledge acquired from colloquial language datasets and the advanced semantic comprehension of LLM, accurately discerns that the user is indeed confirming ongoing business operations. It then adeptly formulates an appropriate follow-up question, verifying the accuracy of the address information.

\begin{figure}[t!]
  \centering
  \begin{subfigure}{0.41\textwidth}
    \includegraphics[width=\textwidth]{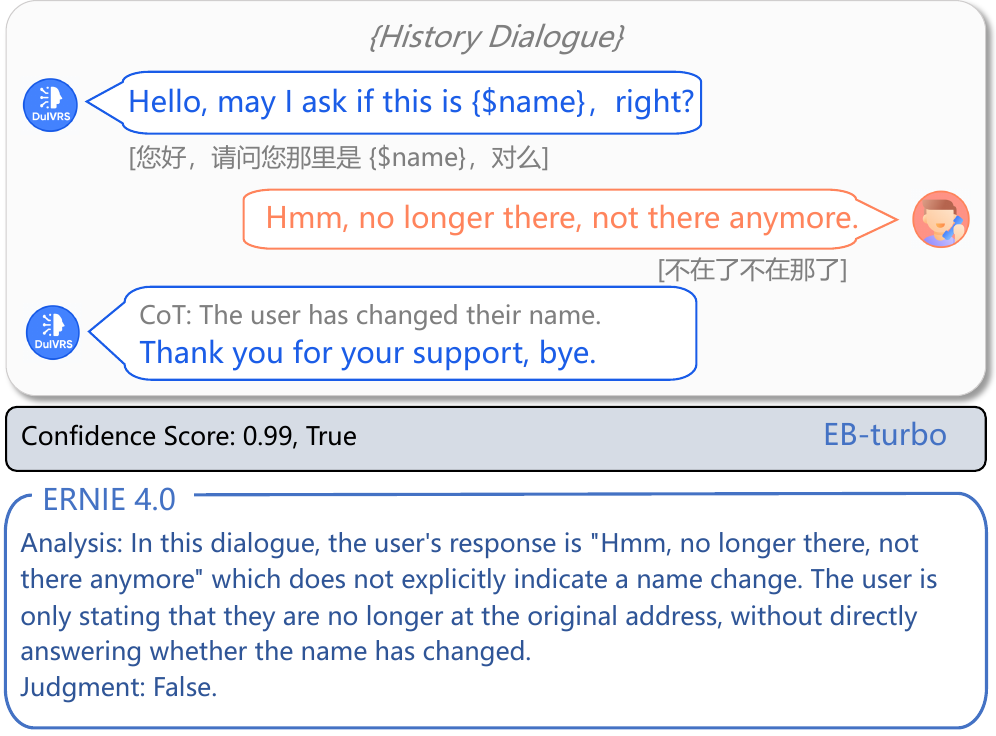}
    \caption{Evaluation for long text understanding.}
    \label{fig:case2-1}
  \end{subfigure}
  \hfill
  \begin{subfigure}{0.41\textwidth}
    \includegraphics[width=\textwidth]{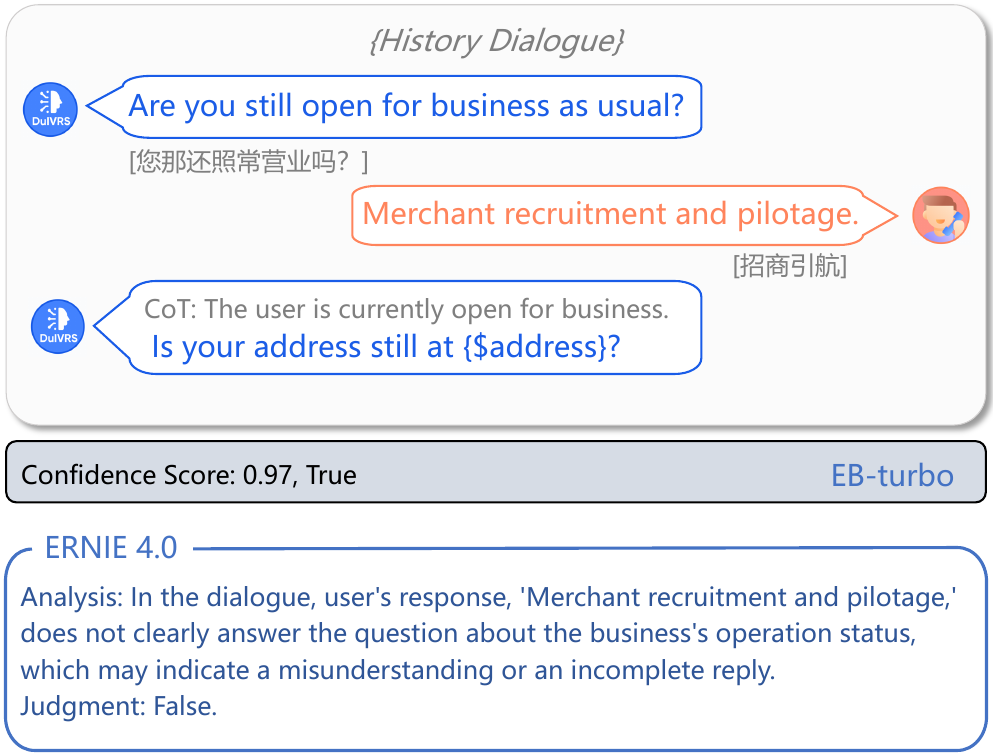}
    \caption{Evaluation for ASR noise.}
    \label{fig:case2-2}
  \end{subfigure}
  % \vspace{-3mm}
  % \caption{Cooperative evaluation of \eb and EB-turbo.}
  \caption{Case studies of cooperative evaluation under challenging scenarios. The two examples show how ERNIE 4.0 and EB-turbo jointly assess response correctness in different settings.}
  \label{fig:case2}
\end{figure}

Figure~\ref{fig:case2} illustrates two cases of how Black-box LLM and LLM-L collaborate during the evaluation phase.
In this process, we combine the superior natural language understanding abilities of ERNIE 4.0 and the specific domain knowledge of the fine-tuned EB-turbo for cooperative evaluation.
Specifically, in the first case, the EB-turbo erroneously interprets the user intention as a name change, assigning a high confidence score of 0.99. In contrast, \eb identifies the error in the agent's inference by leveraging its enhanced semantic understanding capabilities.
The second case presents a scenario where the user response may contain ASR noise.
In Chinese, \Chi{62}{DB}\Chi{55}{46}\Chi{5F}{15}\Chi{82}{2A}$\ $ (Zhao Shang Yin Hang) and \Chi{71}{67}\Chi{5E}{38}\Chi{84}{25}\Chi{4E}{1A}$\ $(Zhao Chang Ying Ye) sound quite similar in the pronunciation of certain regions, so it is hard for \eb to distinguish while EB-turbo can address this ASR noise through fine-tuning. In summary, the cooperative evaluation can take advantage of the two channels to achieve comprehensive evaluation.

\subsection{Illustration of Finite State Machine}
\label{ill_fsm}

To provide an intuitive understanding of the FSM introduced in Section~\ref{subsection:DA}, we provide a visual example that demonstrates both the structural components of the FSM and its real-world instantiation in a dialogue.

\begin{figure}[t!]
\includegraphics[width=0.98\linewidth]{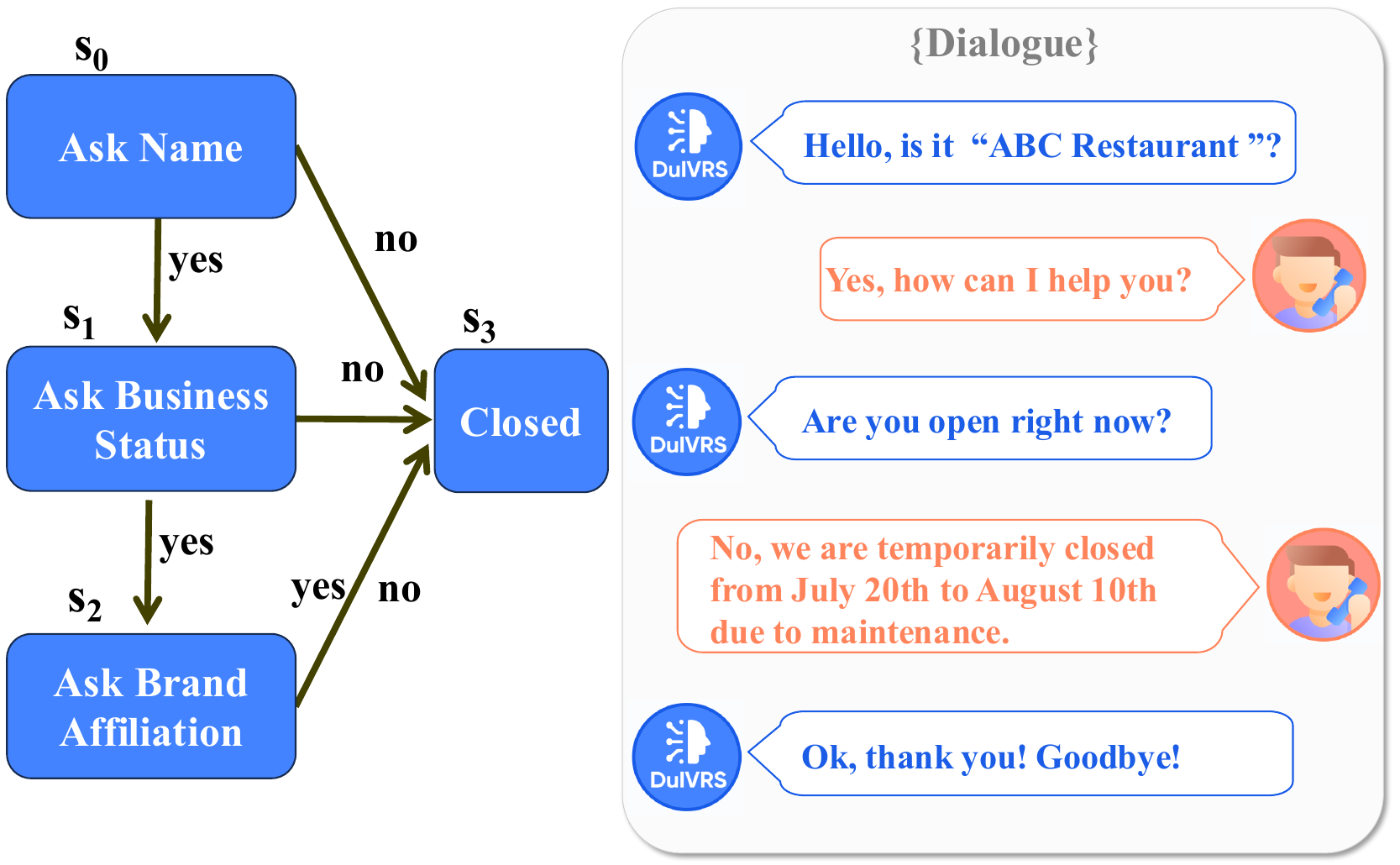} 
\caption{Illustration of the FSM structure (left) and its application in a dialogue interaction (right).}
\label{fig:fsm_illustration}
\end{figure}

As shown on the left of Figure~\ref{fig:fsm_illustration}, the FSM is composed of a finite set of states $\{s_0, s_1, s_2, s_3\}$, where each state corresponds to a specific AI intent such as asking for a POI's name, business status, or brand affiliation. Transitions between states are triggered by user responses (e.g., "yes" or "no"). For example, a transition from $s_0$ to $s_1$ occurs when the user affirms the system’s inquiry about the merchant’s name.

Each complete path from $s_0$ to $s_3$ represents a valid multi-turn dialogue trajectory. For instance:

\begin{itemize}
    \item A full path $s_0 \rightarrow s_1 \rightarrow s_2 \rightarrow s_3$ may reflect a scenario where the AI queries all three attributes before concluding the session.
    \item A shorter path $s_0 \rightarrow s_3$ occurs when the conversation ends early (e.g., merchant is closed).
\end{itemize}

On the right, we show an actual dialogue corresponding to the path $s_0 \rightarrow s_1 \rightarrow s_3$:

\begin{itemize}
    \item At $s_0$, the agent initiates the interaction by confirming the POI name.
    \item At $s_1$, it asks whether the store is open.
    \item The user replies that the store is temporarily closed, prompting a transition to the terminal state $s_3$.
\end{itemize}

This example highlights how FSM-guided generation allows the model to follow well-defined dialogue structures derived from DuIVRS-1 logs. Such structure ensures robust behavior during both training (via data augmentation) and deployment (via constrained decoding). In the context of data augmentation, each valid state transition sequence (e.g., $\{s_0, s_1, s_3\}$ or $\{s_0, s_1, s_2, s_3\}$) is used to synthesize new dialogue samples. During augmentation, transitions are uniformly sampled to avoid over-representing frequent paths in the original logs, promoting a more balanced and comprehensive training distribution.

\begin{figure*}[t]
\centering
\includegraphics[width=0.95\linewidth]{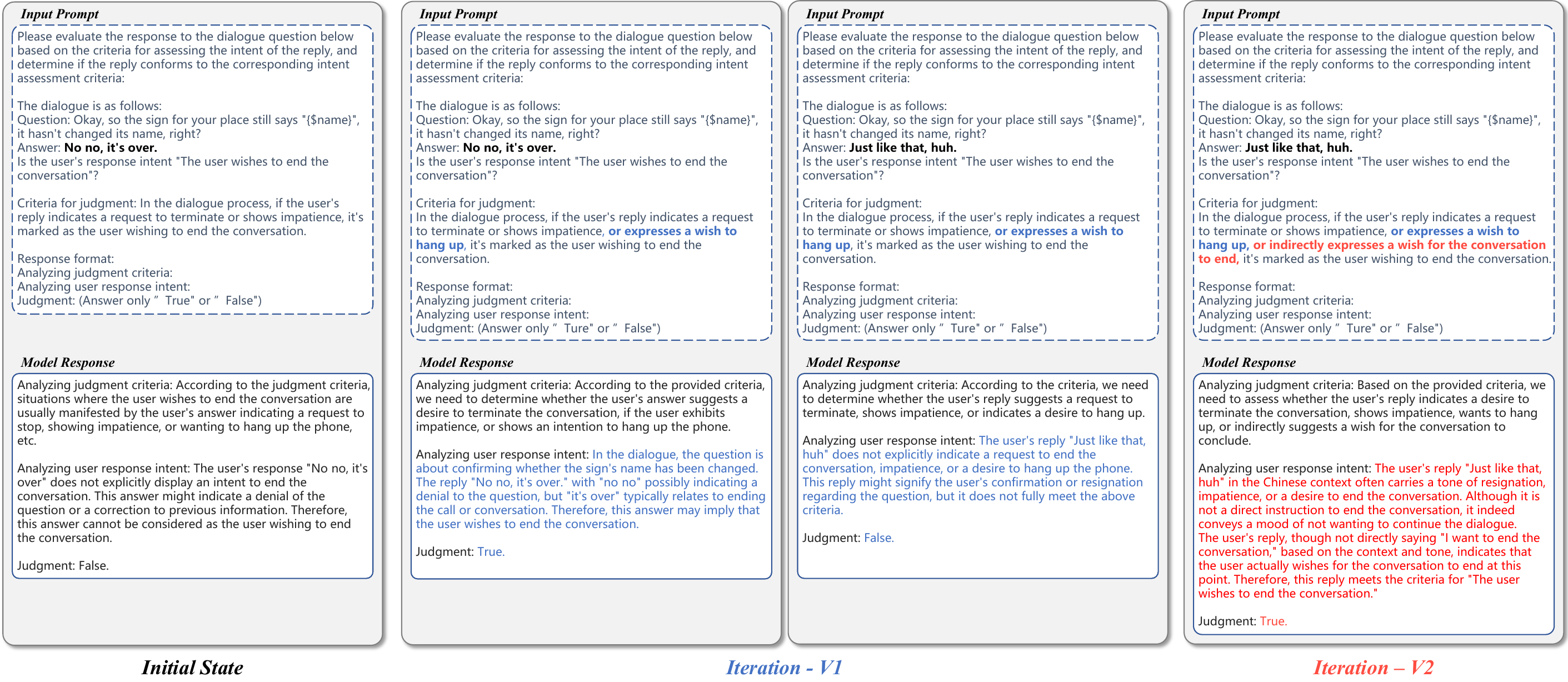} 
\caption{The prompt iteration of \eb in evaluation stage.}
\label{fig:prompt_iteration}
\end{figure*}
% \FloatBarrier

\subsection{Prompt Details}
\label{app:prompt}

In this section, we showcase various prompts devised for both the training and evaluation phases of DuIVRS-2.
The prompt iteration process for \eb (Black-box LLM) during the evaluation phase is illustrated in Figure~\ref{fig:prompt_iteration}. 
Iteration-V1 enhances the prompt by integrating the criterion ``expresses a wish to hang up'', and iteration-v2 further refines it by adding ``indirectly expresses a wish for the conversation to end'', thus extending coverage to more domain-specific edge cases and enhancing the discernment precision of \eb.

\begin{figure}[htb]
  \centering
  \begin{subfigure}{0.425\textwidth}
    \includegraphics[width=1.0\linewidth]{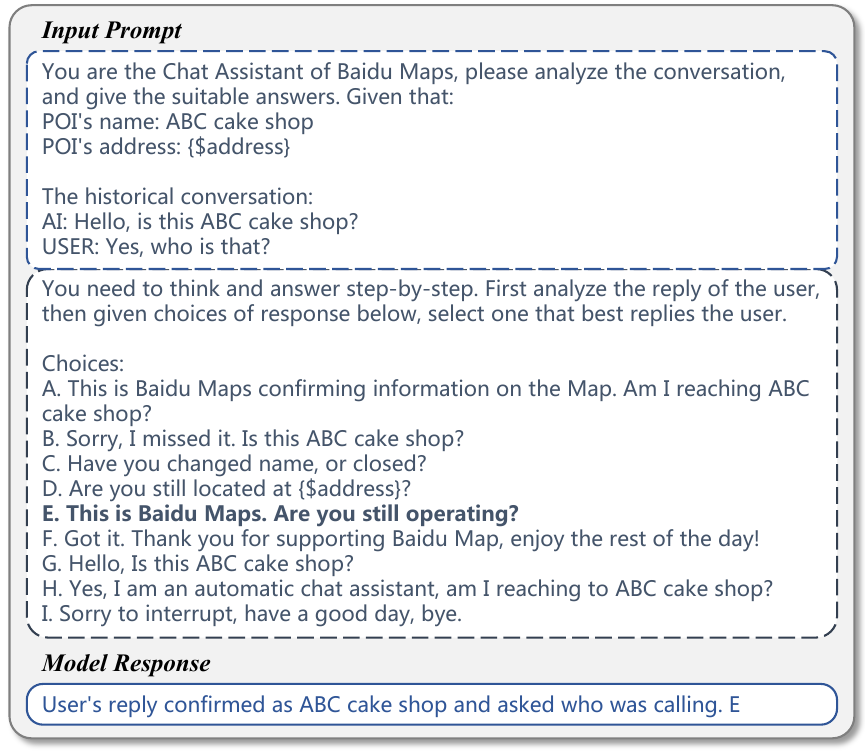} 
    \caption{Example for training the inference model.}
    \label{fig:prompt_generate}
  \end{subfigure}
  \hfill
  \begin{subfigure}{0.425\textwidth}
    \includegraphics[width=1.0\linewidth]{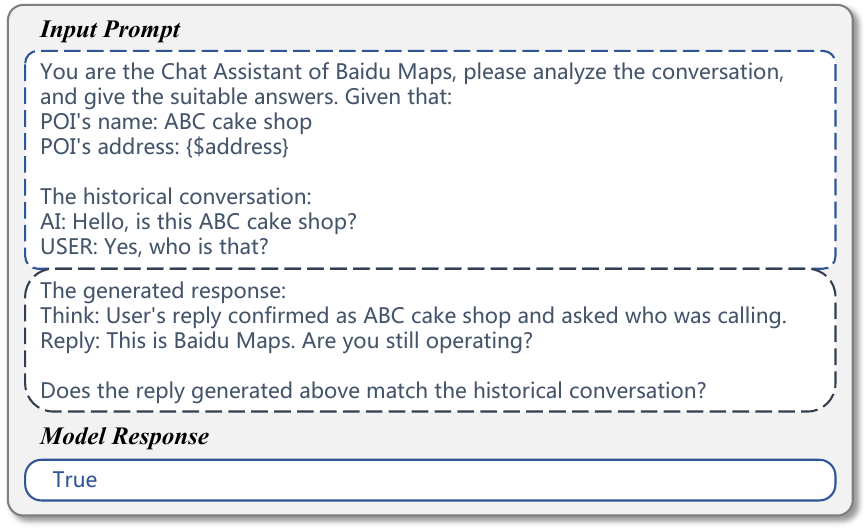} 
    \caption{Example for training the evaluation model.}
    \label{fig:prompt_judge}
  \end{subfigure}
  \caption{Training example for fine-tuning LLMs.}
  \label{fig:prompt_tune}
\end{figure}

Furthermore, Figure~\ref{fig:prompt_generate} provides an example for training the inference model, which adopts a selective generation strategy with CoT mechanism to ensure both safety and stability upon deployment. The model response, ``User's reply confirmed as ABC cake shop'' serves as an instance of the chain of thought process, with option ``E'' indicating the subsequent query ``This is Baidu Maps. Are you still operating?''. 
Figure~\ref{fig:prompt_judge} offers insight into the training evaluation ability of EB-turbo, where it functions to assess the accuracy of EB-tiny's outputs by yielding a ``True/False'' verdict.

\subsection{Illustrative Examples of Hallucination}~\label{sec:hallucination}
To better illustrate the nature of hallucinations observed in ablation variants, we present two representative examples:

\begin{itemize}
    \item \textbf{Direct-SFT (semantic contradiction):}  
     \textit{Context:}  
    \begin{quote}
    Target: Confirm name and business hours. \\
    Model: Is this “Meiyijia Huangcun Branch”? \\
    User: Yes. \\
    \textbf{Model (Hallucinated):} Do you offer parking services?
    \end{quote}
    \textit{Analysis:} The target is to confirm name and business hours, and the model had just confirmed the POI name. The model unexpectedly shifted to inquire about parking availability, which contradicts the  current task and disrupts logical continuity.
    
    \item \textbf{w/o-CoT (FSM constraint violation):}  
\textit{Context:}  
\begin{quote}
FSM-defined options at this turn: \{A. What time do you close?, B. What time do you open?, C. What are your business hours?\} \\
\textbf{Model (Hallucinated):} D
\end{quote}
\textit{Analysis:} At this turn, the system is constrained to selecting among three predefined response options (A–C) generated by the FSM. However, the model outputs option “D,” which is not part of the allowed set.

\end{itemize}

\subsection{Parameter Sensitivity}
As mentioned before, there is a hyperparameter in LLM-L evaluator, namely $\alpha$. In this part, we discuss the parameter sensitivity and identify its optimal value in our evaluation benchmark, which is crucial for assessing the quality of content generated by EB-tiny throughout the iteration process.
We use area under the receiver operator characteristic curve (AUC) from the benchmark for evaluation purposes, as illustrated in Figure~\ref{fig:parameter_sensitivity}. 
We find that the likelihood of content produced by EB-tiny significantly influences the judgment outcomes, pinpointing the best $\alpha$ parameter value at 0.1.

\subsection{Additional Discussion}

\subsubsection{Temporal Scope of This Work}

DuIVRS-2 should be viewed in the temporal context of its development. This project was initiated in 2023, when directly deploying frontier-scale general-purpose LLMs as real-time dialogue managers for large-scale IVR systems was still challenging under strict latency, cost, stability, and controllability constraints. Accordingly, our focus is on designing a deployable LLM-based IVR framework, rather than simply applying the largest available model.

This temporal scope motivates our use of a compact LLM-S for online inference, FSM-constrained selective generation for output stability, and cooperative iterative learning for data and policy refinement. Although recent LLMs have become substantially stronger, the task-specific adaptation and system-level design remain important for long-tail and latency-sensitive POI attribute acquisition. Future work will explore replacing each component with more recent LLMs while maintaining the deployment requirements of industrial IVR systems.

\subsubsection{Discussion on Iterative Learning Risks}
\label{iter_risk}

While the cooperative iterative learning framework introduced in DuIVRS-2 demonstrates significant performance improvements across iterations, we also observed several potential risks inherent in iterative processes:

\textbf{Overfitting Risk:}
Continuous iterations on progressively refined datasets may inadvertently introduce overfitting, particularly if the iterative refinement overly emphasizes specific frequent or well-represented scenarios, thus impairing the generalization capability. To mitigate this risk, we intentionally integrated diversified synthetic data generation via FSM-based augmentation and maintained a rigorous evaluation process combining both fine-tuned LLM (EB-turbo) and a domain-agnostic black-box LLM (ERNIE 4.0), thereby preserving generalization and robustness.

\textbf{Diminishing Returns:}
As illustrated in Figure~4a, iterative refinement eventually encounters diminishing returns, with performance gains becoming increasingly marginal after several iterations. Empirical results suggest that optimal performance is typically achieved around the third or fourth iteration. Beyond this point, the cost (computational resources and human annotation efforts) associated with additional iterations may outweigh marginal gains in performance.

\textbf{Error Propagation and Amplification:}
Incorrect annotations or evaluation inaccuracies during early iterations can propagate through subsequent iterations, potentially amplifying errors and adversely affecting system reliability. We address this concern by leveraging the cooperative voting mechanism between domain-specific and domain-agnostic LLM evaluators, coupled with periodic human judgment interventions, effectively curbing error propagation risks.

To effectively manage these iterative learning risks, we recommend systematically monitoring performance metrics across iterations, maintaining balanced data augmentation, and establishing clear criteria for iteration termination based on performance saturation and cost-effectiveness analyses.

\subsubsection{Discussion on ASR Errors}
\label{asr_error}

Automatic Speech Recognition (ASR) errors represent a significant challenge for interactive voice response systems like DuIVRS-2, particularly when dealing with diverse dialects, accents, and variable speech quality. Although ASR optimization was not the primary focus of our study, our approach integrated dialect-inclusive corpora during ASR training to enhance recognition robustness across regions. Furthermore, our dialogue management strategy deliberately employs concise and structured query phrasing, prompting user replies primarily as short, clear responses (e.g., "Yes" or "No"), thereby minimizing ambiguity and dialect-induced errors.

The iterative learning framework implemented in DuIVRS-2 incorporates continuous feedback from real-world user interactions, progressively mitigating the impact of ASR inaccuracies. As illustrated in Figure 7(b) in Section 4.6, common ASR-induced errors include phonetic confusion and misrecognition of acoustically similar phrases. Such errors could significantly affect the interpretation of user intent, thus influencing subsequent conversational decisions made by the system.

To systematically address these challenges, DuIVRS-2 employs two main strategies: first, an FSM-based structured dialogue design limits potential confusion stemming from ASR errors; second, iterative refinements and continuous human-in-the-loop evaluations further minimize error propagation. Our ongoing analyses suggest that these strategies substantially reduce ASR-related misinterpretations, ensuring stable and robust interaction outcomes even in complex acoustic environments.

\end{document}